\documentclass[10pt]{article}
\usepackage[preprint]{tmlr}

\usepackage{amsmath,amsfonts,bm}

\def\eqref#1{equation~\ref{#1}}

\def\1{\bm{1}}

\DeclareMathAlphabet{\mathsfit}{\encodingdefault}{\sfdefault}{m}{sl}
\SetMathAlphabet{\mathsfit}{bold}{\encodingdefault}{\sfdefault}{bx}{n}

\usepackage{url}
\usepackage{graphicx}
\usepackage{booktabs}
\usepackage{array}
\usepackage{placeins}
\usepackage{float}
\usepackage{needspace}
\usepackage[pagebackref=false, breaklinks=true, colorlinks, citecolor=blue, bookmarks=false]{hyperref}

\newcommand{\tablefont}{\footnotesize}

\title{Backbone-Informed Adaptation after EEG Pretraining:\\
A Capacity-Controlled Study}

\author{%
  \makebox[\textwidth][c]{%
    \begin{tabular}{@{}*{3}{>{\centering\arraybackslash}p{0.30\textwidth}}@{}}
      \textbf{Mingyang Jiang}$^{1}$ &
      \textbf{Yamin Li}$^{1}$ &
      \textbf{Fan Ma}$^{2}$ \\[0.6em]
      \textbf{Daniel Moyer}$^{1}$ &
      \textbf{Hua Xu}$^{2}$ &
      \textbf{Catie Chang}$^{1}$ \\[1em]
      \multicolumn{3}{c}{\normalfont\small\itshape
        $^{1}$ Department of Computer Science, Vanderbilt University} \\
      \multicolumn{3}{c}{\normalfont\small\itshape
        $^{2}$ Department of Biomedical Informatics and Data Science, Yale School of Medicine}
    \end{tabular}%
  }%
}

\begin{document}

\maketitle

\begin{abstract}
Electroencephalography (EEG) foundation models learn representations that
combine information across channels and time. This structure motivates
adapters that model the same relationships during downstream learning. 
However, it is not clear whether such a design outperforms other candidate adapters. Moreover, in order to arrive at a fair comparison between candidate adapters,
we also need to consider their trainable parameters.
Here, we 
ask whether a backbone-informed (\textit{native})
adapter makes better use of a given capacity by comparing it against a token-wise multilayer perceptron control that transforms
each feature vector independently. We approximately match their trainable
parameter counts and keep the pretrained encoder fixed. Across two EEG
foundation models and tasks covering emotion recognition, sleep staging, and
EEG event classification, the native adapter shows no consistent advantage.
For one backbone, differences between the designs are small relative to
variation across runs. For the other, the token-wise control performs better
on emotion recognition and EEG event classification. On sleep staging, the
native adapter's advantage over the token-wise control is reduced when we also train the component that
processes sequences of EEG segments. The relative performance of the
two adapters depends on the backbone, task, and downstream training protocol.
These findings
motivate evaluating backbone-informed adapters against capacity-matched
alternatives.\footnote{The authors used a large language model to assist with drafting
and editing. The authors remain responsible for the manuscript's scientific
content, analyses, and claims.}
\end{abstract}

\begin{quote}
\noindent\textbf{Code availability:}
\url{https://github.com/0oJimmyo0/EEGxPlore-TMLR}
\end{quote}

\section{Introduction}
\label{sec:introduction}

Electroencephalography (EEG) records brain electrical activity across channels
and time, supporting decoding tasks such as emotion recognition, sleep staging, and EEG event
classification. Self-supervised EEG foundation models learn
representations from unlabeled recordings for reuse across downstream tasks \citep{kostas2021bendr,jiang2024large,wang2025cbramod}. Because pretraining
is not tailored to every downstream task, these features may still need
task-specific adjustments. Downstream adaptation either updates pretrained
weights through full fine-tuning or trains fewer task-specific parameters
using parameter-efficient methods \citep{lee2025brainwave,jeon2025tasteful}.
We focus on adapters that operate after a fixed pretrained encoder. In this setting, the way in which pretrained features are transformed is a central design choice.

Existing EEG adaptation methods have explored different ways to modify
pretrained representations, including emphasizing task-relevant features
\citep{jeon2025tasteful}, modeling sensor relationships
\citep{suzumura2024graphadapter}, and using class prototypes \citep{ma2026scope}.
Another design choice is whether an adapter should explicitly follow the
channel and temporal organization of the pretrained encoder. Mixing information
across channels and short time segments, or \emph{temporal patches}, may help
adapt these representations to downstream tasks, but the encoder already
models relationships across these dimensions. It is therefore unclear whether
an adapter that follows this organization is preferable to one that transforms
each feature vector independently. Improvements over a frozen-feature classifier
alone cannot resolve this question, because adding an adapter also introduces
trainable parameters. This motivates comparing the two designs at similar
trainable capacity.

\begin{figure}[!htbp]
\centering
\includegraphics[width=\linewidth]{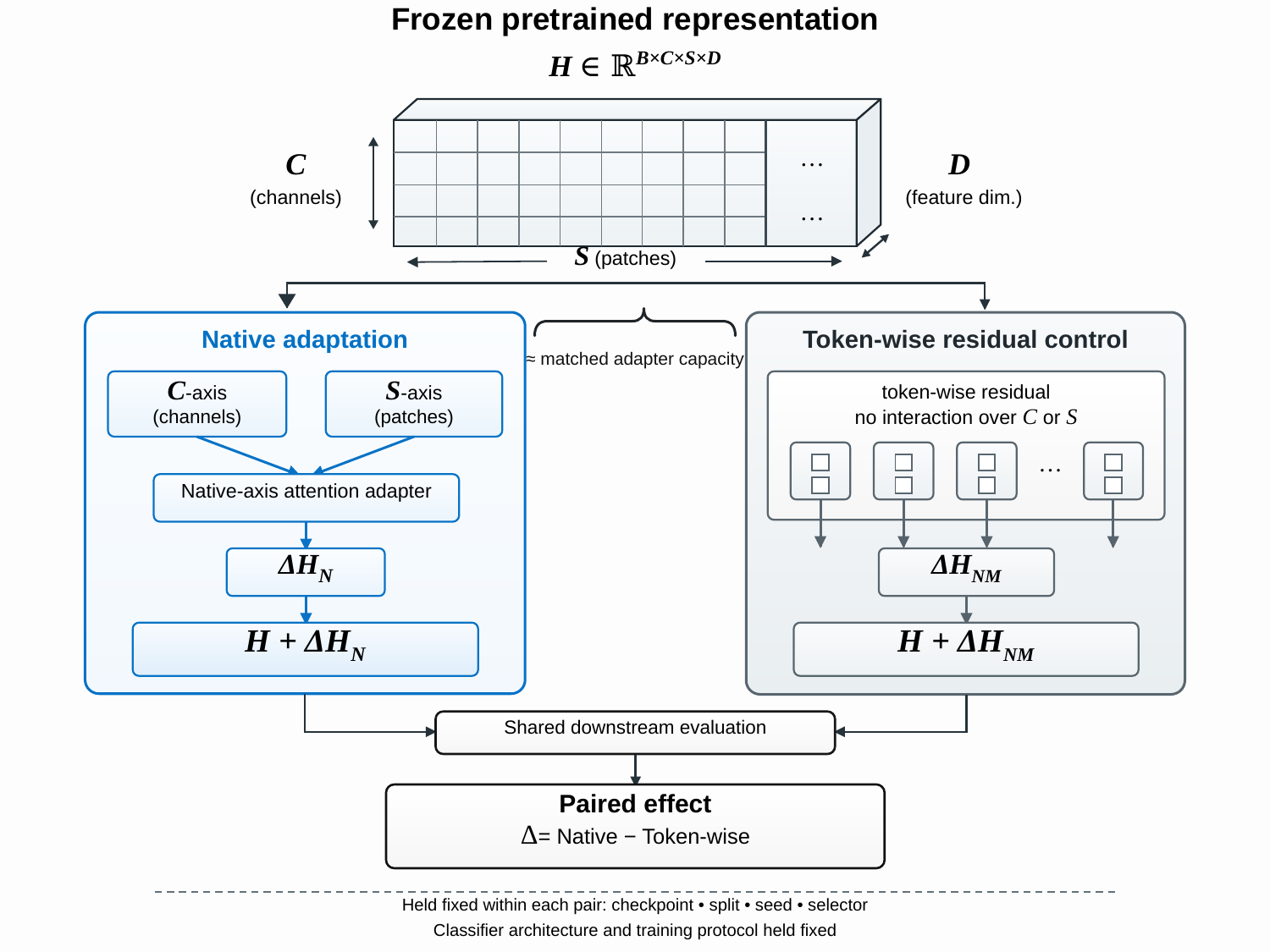}
\caption{Primary controlled comparison. A common pretrained representation is
passed to an additional task-supervised native-axis attention adapter and a
parameter-matched token-wise control. Within each pair, checkpoint,
split, seed, classifier architecture and training protocol, and validation
rule are held fixed; the paired effect is native minus
token-wise performance.}
\label{fig:study-overview}
\end{figure}

To examine this question, we compare two designs that receive the same
pretrained features and precede the same classifier architecture
(Figure~\ref{fig:study-overview}). Our backbone-informed design is a
\emph{native-axis attention adapter} that mixes information along the channel
and temporal-patch axes of the pretrained representation. The token-wise
multilayer perceptron (MLP) control instead transforms each feature vector,
or \emph{token}, independently. We approximately match their trainable adapter
parameter counts and keep the pretrained encoder fixed. This comparison
evaluates the complete designs, including differences in their operators and
token-mixing behavior.

We  conduct our study with two representative EEG foundation models, CBraMod and LaBraM. CBraMod applies channel and temporal
attention to separate feature subspaces, while LaBraM produces patch tokens
that can be arranged into a channel-by-patch grid
\citep{jiang2024large,wang2025cbramod}, providing two feature organizations for
constructing the adapters. The primary comparison (RQ1) asks whether the native
adapter outperforms the capacity-matched token-wise control. RQ2 provides
context by asking whether the native adapter improves over the
\emph{frozen-backbone probe}, a classifier trained on the same fixed features
without an adapter. RQ3 is a secondary analysis of whether the native adapter
adds value when the backbone itself is trainable.

Under the evaluated protocols, CBraMod differences are small relative to seed
variation. LaBraM favors the token-wise
control on emotion recognition (FACED) and EEG event classification (TUEV).
On sleep staging (ISRUC), the native adapter's balanced-accuracy advantage
over the token-wise control shrinks when we train the downstream sequence
encoder, which combines information across consecutive EEG segments
(Section~\ref{sec:boundary-sequence}). A branch-preserving MLP performs similarly
to the token-wise control across the LaBraM comparisons.
We also compare full fine-tuning with and without the native adapter across
four settings: CBraMod on FACED, TUEV, and SEED-V, and LaBraM on PhysioNet-MI.
Adding the adapter yields no consistent benefit.
These results motivate comparing backbone-informed adapters with
parameter-matched alternatives.

\section{Related Work}
\label{sec:related-work}

\subsection{EEG representations and feature organization}

The structure of a pretrained EEG encoder determines the features and organization available to a
downstream adapter. Self-supervised models learn reusable representations
through different pretraining approaches
\citep{kostas2021bendr,yang2023biot,mohammadi2024eeg2rep,yue2024braingpt}.
Electrode-location encoding, retrieval-based spatial learning, and alternating
spatial--temporal attention organize spatial and temporal information
\citep{foumani2025eegx,ding2025brainpro,dimofte2025cerebro}, while
in frequency-balanced pretraining, the learned
representation is informed by the frequency content \citep{yu2026fame}. Our study asks how to adapt these features
once pretraining is complete. LaBraM's patch representations and CBraMod's
criss-cross channel--temporal attention
\citep{jiang2024large,wang2025cbramod} provide two feature organizations for
constructing and comparing post-backbone adapters.

\subsection{Adaptation designs and structural priors}

Parameter-efficient methods offer different ways to modify a pretrained
network. Bottleneck adapters add down-projection, nonlinearity, and
up-projection modules within the network \citep{houlsby2019parameter}, while
LoRA learns low-rank weight updates \citep{hu2022lora}. Localized LoRA restricts
structured updates to selected weight-matrix blocks
\citep{barazandeh2025localizedlora}. For EEG foundation models,
\citet{lee2025brainwave} compare full fine-tuning with parameter-efficient
alternatives, including LoRA. These methods provide broader tuning references
for our comparison, which fixes adapter placement after the backbone.

A separate design choice concerns what guides adaptation.
\citet{jeon2025tasteful} focus pretrained EEG features on information relevant
to downstream tasks. \citet{suzumura2024graphadapter} use a graph-based adapter
to learn spatial relationships among EEG sensors while keeping a pretrained
temporal backbone fixed. \citet{ma2026scope} guide lightweight adaptation
with structured class prototypes under limited supervision. These methods
use task, sensor, or supervision information to guide adaptation. Our question
is whether following the backbone's channel and temporal-patch axes improves
performance over a token-wise alternative with similar adapter capacity.

Token mixing need not use attention: MLP-Mixer uses cross-token MLPs, while
the PoolFormer instantiation of MetaFormer uses pooling \citep{tolstikhin2021mlpmixer,yu2022metaformer}.
Our token-wise control transforms tokens independently. The comparison
evaluates complete adapter designs rather than isolating the mixer operator.

\subsection{Evaluation protocols and controlled comparisons}

The relative performance of different adapters needs to be interpreted in the context of the evaluation
protocol. Literature on EEG foundation models indicates that models, preprocessing, data splitting, the unit of
evaluation, and probing versus fine-tuning are relevant choices
\citep{kuruppu2025critical,liu2026eegfmbenchmark}, and related neuroimaging work
shows that preprocessing can affect prediction and representation similarity
\citep{li2023pipeline}. Further, evaluation
across tasks, datasets, and distribution shifts provides a more complete picture of generalizability
\citep{xiong2025eegfmb,banville2026neuralbench,lee2026neuroadapt}.
For the adapter comparison, our study holds the checkpoint, split, classifier
protocol, and training recipe fixed within each primary pair.
We select each training run’s checkpoint using only validation performance, and compare the adapters’ held-out test results from runs with the same random seed.

\section{Adaptation Framework}
\label{sec:adaptation-framework}

The native adapter and token-wise control add residual transformations to the
post-backbone representation. CBraMod's native branches use separate feature
halves, whereas LaBraM's branches use the full token embedding. Both adapters
preserve the feature-grid shape and feed the same classifier architecture.

\subsection{Representation format}
\label{sec:representation-geometry}

We denote the post-backbone feature grid by
\begin{equation}
  H \in \mathbb{R}^{B \times C \times S \times D},
\end{equation}
where $B$ is the batch size, $C$ is the channel dimension, $S$ is the
temporal-patch dimension, and $D$ is the feature dimension. The adapter
preserves this shape. It constructs axis-specific sequences from the grid,
applies an interaction mixer, and reshapes the output to the original
format.

\begin{figure}[t]
\centering
\includegraphics[width=\linewidth]{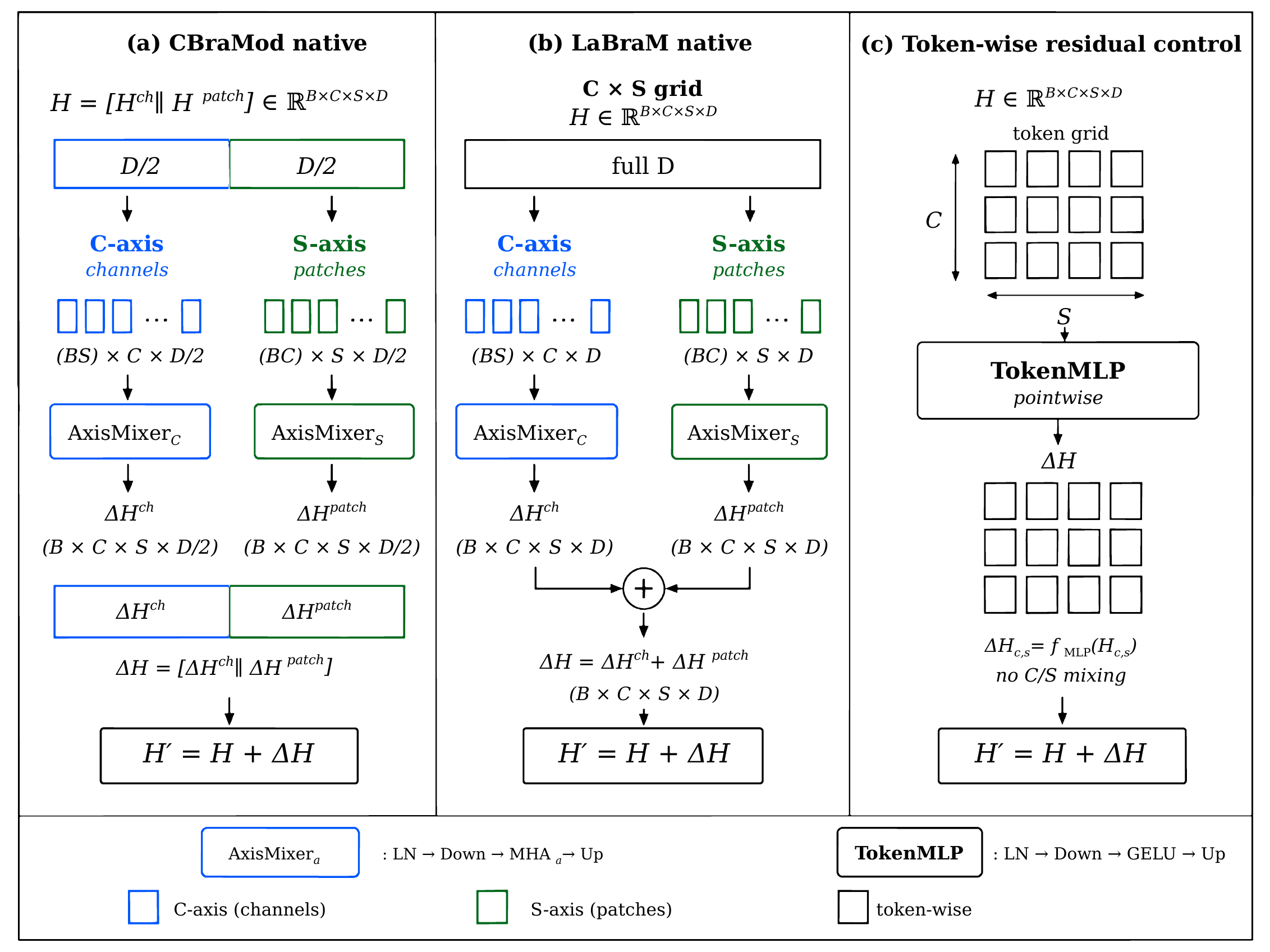}
\caption{Backbone-specific native adapters and the token-wise control
used in RQ1. CBraMod and LaBraM follow their
respective representation formats; the token-wise control applies
an independent transformation without channel- or patch-axis mixing.
Section~\ref{sec:parameter-matching} reports adaptation-module parameter
counts and matching tolerances.}
\label{fig:adapter-architecture}
\end{figure}

\subsubsection{CBraMod format}

CBraMod uses parallel channel and temporal attention \citep{wang2025cbramod}.
The implementation used here applies channel attention to the first $D/2$
features and temporal attention to the second $D/2$. Our native adapter follows
this branch partition:
\begin{equation}
  H = [H^{\mathrm{ch}} \mathbin{\|} H^{\mathrm{patch}}],
  \qquad
  H^{\mathrm{ch}},H^{\mathrm{patch}}
  \in \mathbb{R}^{B \times C \times S \times D/2}.
\end{equation}

The channel branch reshapes each temporal patch into a sequence over channels,
\begin{equation}
  H^{\mathrm{ch}}
  \longrightarrow
  \mathbb{R}^{(BS) \times C \times D/2},
\end{equation}
whereas the patch branch reshapes each channel into a sequence over temporal
patches,
\begin{equation}
  H^{\mathrm{patch}}
  \longrightarrow
  \mathbb{R}^{(BC) \times S \times D/2}.
\end{equation}
The channel-only and patch-only variants enable the corresponding branch; the
channel-plus-patch variant enables both branches independently.

\subsubsection{LaBraM format}

LaBraM produces a sequence of patch tokens that can be reorganized into a
channel-by-patch grid \citep{jiang2024large}. Its native adapter
uses the full embedding dimension. The
channel and patch branches construct sequences as
\begin{equation}
  H \longrightarrow \mathbb{R}^{(BS) \times C \times D}
  \qquad\text{or}\qquad
  H \longrightarrow \mathbb{R}^{(BC) \times S \times D},
\end{equation}
respectively. The primary LaBraM runs use low-rank attention branches;
Appendix~\ref{app:primary-configs} records the architecture and initialization
settings.

\subsection{Native attention branch}
\label{sec:residual-adapter}

For each axis included in the adapter, the native attention branch extracts the corresponding sequence
from $H$, applies a bottleneck attention block along that sequence, and places
the resulting update back into the original feature grid. For an enabled
interaction axis $a$, let $\mathcal{R}_a$ denote the associated
residual branch. Let $\Phi_a$ denote the backbone-specific extraction operator
that selects the relevant feature subspace and forms the sequence for axis $a$,
and let $\Psi_a$ denote the corresponding restoration operator. For CBraMod,
$\Phi_a$ selects the corresponding $D/2$ subspace; for LaBraM, it operates on
the full $D$-dimensional embedding. The branch applies layer
normalization, projects the features to a bottleneck dimension, performs
self-attention over the selected axis, and projects the result back:
\begin{align}
  X_a &= \Phi_a(H), \\
  Z_a &= \operatorname{Down}_a\!\left(\operatorname{LN}(X_a)\right), \\
  U_a &= \operatorname{MHA}_a(Z_a,Z_a,Z_a), \\
  \mathcal{R}_a(H) &= \Psi_a\!\left(\operatorname{Up}_a(U_a)\right).
\end{align}
The adapted representation is
\begin{equation}
  H_{\mathrm{adapted}}
  = H + \sum_{a \in \mathcal{A}} \lambda_a\mathcal{R}_a(H),
  \label{eq:adapter-residual}
\end{equation}
where $\mathcal{A}$ is the set of enabled axes and $\lambda_a$ denotes the
effective coefficient induced by the learned branch scale and configured
residual multiplier. For the CBraMod primary configurations, the output
projections are zero-initialized, so the residual update is initially zero and
$H_{\mathrm{adapted}}=H$ at initialization. The primary LaBraM configuration
uses a small learned initial branch scale without zero-initializing the output
projection; its initialization settings are reported in
Appendix~\ref{app:primary-configs}. The bottleneck reduces trainable capacity
relative to updating the backbone.

\subsection{Token-wise control}
\label{sec:native-token-wise}

The token-wise control applies an independent bottleneck
transformation to each channel--patch token:
\begin{equation}
  \mathcal{R}_{\mathrm{tw}}(H)
  = \operatorname{Up}\!\left(
      \operatorname{GELU}\!\left(
        \operatorname{Down}\!\left(\operatorname{LN}(H)\right)
      \right)
    \right).
  \label{eq:token-wise}
\end{equation}
We adjust the token-wise bottleneck width to match the native adapter's
trainable parameter count within the tolerance in
Section~\ref{sec:parameter-matching}. RQ1 compares these complete attention
and MLP designs, so operator choice and axis mixing vary together.

\section{Experimental Protocol}
\label{sec:experimental-protocol}

RQ1 tests whether the native adapter outperforms the capacity-matched token-wise
control. RQ2 tests its benefit over a frozen-backbone probe, and RQ3 its
incremental benefit during full fine-tuning.

\subsection{Datasets}
\label{sec:datasets-evidence-roles}
We conduct our analysis on five public downstream datasets (Table~\ref{tab:dataset-roles}): FACED
(emotion recognition)~\citep{chen2023faced}, ISRUC (sleep staging)
\citep{khalighi2016isruc}, TUEV (EEG event classification)
\citep{obeid2016tuh,harati2015tuev}, PhysioNet-MI (motor imagery)
\citep{goldberger2000physiobank,schalk2009eegmmidb}, and SEED-V (emotion recognition)
\citep{liu2022seedv}.

\begin{table}[!ht]
\centering
\caption{Datasets and frozen-backbone comparisons. RQ1 compares the native
adapter with the capacity-matched token-wise control; RQ2 compares it with
the frozen-backbone probe. $C$ denotes channels and $S$ temporal patches.
TUEV record/prefix splits do not establish subject disjointness.}
\label{tab:dataset-roles}
\vspace{\baselineskip}
\tablefont
\setlength{\tabcolsep}{3pt}
\begin{tabular}{@{}llrrrll@{}}
\hline
\textbf{Dataset} & \textbf{Task} & \textbf{$C$} & \textbf{$S$} & \textbf{Classes} & \textbf{Comparisons} & \textbf{Split unit} \\
\hline
FACED & Emotion & 32 & 10 & 9 & RQ1, RQ2 & Subject/source file \\
ISRUC & Sleep staging & 6 & 30 & 5 & RQ1, RQ2 & Subject \\
TUEV & Event classification & 16 & 5 & 6 & RQ1, RQ2 & Record/prefix \\
PhysioNet-MI & Motor imagery & 64 & 4 & 4 & RQ2 & Subject \\
SEED-V & Emotion & 62 & 1 & 5 & RQ2 (CBraMod) & Within-subject trial \\
\hline
\end{tabular}
\end{table}

FACED, ISRUC, and TUEV each provide multiple channels and temporal patches,
so the native adapter can mix features along both axes. For each dataset and
backbone, we evaluate the native adapter, capacity-matched token-wise control,
and frozen-backbone probe across three seeds. These six settings support
both the adapter-design comparison (RQ1) and the probe comparison (RQ2).

PhysioNet-MI extends RQ2 to motor imagery for both backbones. It has native
adapter and probe runs but no capacity-matched token-wise pair, so it does
not support RQ1. SEED-V has only one temporal patch per example ($S=1$),
leaving no cross-patch interaction to model. Its CBraMod channel-only adapter
and probe runs test whether adaptation helps when only channel mixing is
available. The full-fine-tuning comparisons (RQ3) cover CBraMod on FACED,
TUEV, and SEED-V, and LaBraM on PhysioNet-MI.

We follow the CBraMod data-splitting protocol \citep{wang2025cbramod} for all downstream datasets.
Specifically, FACED, ISRUC, and PhysioNet-MI use subject-disjoint splits; SEED-V uses
within-subject trial splits. TUEV uses record/prefix splits with unverified
subject disjointness. Appendix~\ref{app:preprocessing} gives split counts
and preprocessing details.
Note that the documentation for the released CBraMod weights does not specify whether TUEV was excluded from the TUEG pretraining data \citep{wang2025cbramod}. We therefore do not interpret CBraMod's TUEV results
as evidence of transfer to a dataset unseen during pretraining.

\subsection{Conditions and evaluation}
\label{sec:backbones-conditions}
\label{sec:trainability-regimes}

For the frozen-backbone comparisons, we train the classifier while keeping
the pretrained encoder fixed. The probe uses no adapter, whereas the native
adapter and token-wise control are trained alongside the classifier.
For broader performance context, we include a
generic bottleneck adapter and LoRA, which learns low-rank updates to backbone
weights. We also evaluate upper-layer tuning, which updates the final backbone
layers, and dense fine-tuning, which updates the full backbone
(see details in Appendix~\ref{app:supporting-results}).

Following the released CBraMod implementation \citep{wang2025cbramod}, we use
flattened-feature MLP heads for FACED/TUEV and a projection and Transformer
head for ISRUC. For LaBraM, we follow its mean-pooling and linear-classifier
design \citep{jiang2024large} and add a sequence encoder before the classifier
for ISRUC. Table~\ref{tab:primary-heads} reports the head architectures and
trainable parameter counts.

For ISRUC, both primary protocols freeze the pretrained epoch encoder.
CBraMod trains the downstream sequence encoder within its classifier head.
The primary LaBraM--ISRUC runs freeze a newly initialized downstream
sequence encoder. A separate follow-up
trains the sequence encoder and classifier in all four conditions while
retaining the frozen epoch encoder (Section~\ref{sec:boundary-sequence}).
Appendix~\ref{app:primary-configs} gives the primary configurations.

We run each primary experimental condition with three training seeds. For each run,
we select the checkpoint with the highest Cohen's \(\kappa\) on the validation
set and report its performance on the held-out test set. We use validation
balanced accuracy for selector sensitivity. Held-out test balanced accuracy and macro-F1 are
primary metrics, with test kappa and weighted-F1 secondary.

\subsection{Capacity control and paired comparisons}
\label{sec:parameter-matching}
\label{sec:geometry-eligibility}
\label{sec:paired-analysis}

Adding adapters introduces both a feature transformation and additional trainable parameters. Therefore, performance gains over a frozen-backbone probe may reflect both factors and do not by themselves establish
whether native-axis attention is preferable to a token-wise transformation.
RQ1 compares these designs with closely matched parameter counts, using
the same frozen pretrained features and downstream classifier architecture.

We fix the native attention bottleneck dimension at 64 and adjust the
token-wise bottleneck width to match its parameter count: 148 for CBraMod
and 213 for LaBraM. Matching covers the complete adaptation module,
including projections, normalization, and learned residual scales.
We exclude the downstream head because its architecture and trainability
are identical within each pair. Let \(P_{\mathrm{native}}\) and
\(P_{\mathrm{tw}}\) denote the two adapter parameter counts. We choose widths
with a relative mismatch of at most five percent:
\begin{equation}
  \delta_P
  = \frac{\left|P_{\mathrm{native}}-P_{\mathrm{tw}}\right|}
  {P_{\mathrm{native}}}
  \leq 0.05.
  \label{eq:parameter-mismatch}
\end{equation}
The resulting counts are 59,610 versus 59,949 for CBraMod and 85,810 versus
86,014 for LaBraM, giving mismatches of 0.569\% and 0.238\%, respectively.
These counts apply to each backbone's FACED, ISRUC, and TUEV comparisons.

Within each comparison, we use the same residual multiplier, adapter dropout,
output-initialization policy, optimizer, learning-rate schedule, and
classifier protocol (Appendix~\ref{app:primary-configs}). We use one training
recipe per setting without separate tuning for each adapter.
We also replace attention with token-wise MLPs within the native branches,
approximately matching parameter counts. This tests whether the replacement
changes performance relative to the token-wise control
(Section~\ref{sec:boundary-operator}).

For each metric and seed, we subtract condition \(B\)'s test score from
condition \(A\)'s, comparing runs with the same backbone, dataset, and training
regime. We report the mean and standard deviation of these paired differences
across the three seeds, together with the number of seeds for which \(A\)
outperforms \(B\). These summaries describe the three observed reruns, without
population-level significance claims. Aggregates across settings are descriptive
and do not treat the settings as independent samples.

\FloatBarrier

\section{Results}
\label{sec:results}

\subsection{Primary matched-adapter comparison (RQ1)}
\label{sec:results-primary-effects}

Table~\ref{tab:primary-effects} reports test balanced accuracy and macro-F1
for the six primary comparisons.

\begin{table}[!htbp]
\centering
\caption{Primary held-out test performance and paired effects.
Entries are mean $\pm$ standard
deviation over three seeds. RQ1 is native $-$ token-wise and RQ2 is
native $-$ probe; positive values favor native.
$\dagger$ LaBraM--ISRUC uses the frozen-sequence protocol.
$\ddagger$ LaBraM--TUEV RQ2 effects use a common probe
reference (Appendix~\ref{app:rq2-seed-effects}).}
\label{tab:primary-effects}
\vspace{\baselineskip}
\tablefont
\setlength{\tabcolsep}{2pt}
\begin{tabular}{@{}llccccc@{}}
\toprule
\multicolumn{7}{c}{\textbf{(a) Balanced accuracy}}\\
\textbf{Backbone} & \textbf{Dataset} & \textbf{Probe} & \textbf{Native} &
\textbf{Token-wise} & \textbf{RQ1 $\Delta$} & \textbf{RQ2 $\Delta$} \\
\midrule
CBraMod & FACED  & $.3777\pm.0023$ & $.3910\pm.0039$ & $.3879\pm.0031$ &
$+.0030\pm.0045$ & $+.0133\pm.0061$ \\
 & ISRUC  & $.7562\pm.0033$ & $.7519\pm.0065$ & $.7522\pm.0024$ &
$-.0003\pm.0052$ & $-.0043\pm.0050$ \\
 & TUEV  & $.4656\pm.0315$ & $.4548\pm.0223$ & $.4580\pm.0083$ &
$-.0032\pm.0194$ & $-.0108\pm.0511$ \\
\midrule
LaBraM & FACED  & $.1589\pm.0003$ & $.1613\pm.0097$ & $.1951\pm.0028$ &
$-.0338\pm.0116$ & $+.0023\pm.0094$ \\
 & ISRUC$^{\dagger}$  & $.6515\pm.0021$ & $.7453\pm.0003$ & $.7114\pm.0044$ &
$+.0340\pm.0047$ & $+.0938\pm.0018$ \\
 & TUEV  & $.3658^{\ddagger}$ & $.3916\pm.0182$ & $.4307\pm.0010$ &
$-.0391\pm.0174$ & $+.0258\pm.0182$ \\
\bottomrule
\end{tabular}
\par\vspace{4pt}
\begin{tabular}{@{}llccccc@{}}
\toprule
\multicolumn{7}{c}{\textbf{(b) Macro-F1}}\\
\textbf{Backbone} & \textbf{Dataset} & \textbf{Probe} & \textbf{Native} &
\textbf{Token-wise} & \textbf{RQ1 $\Delta$} & \textbf{RQ2 $\Delta$} \\
\midrule
CBraMod & FACED  & $.3789\pm.0024$ & $.3907\pm.0038$ & $.3874\pm.0015$ &
$+.0033\pm.0041$ & $+.0118\pm.0061$ \\
 & ISRUC  & $.7516\pm.0055$ & $.7485\pm.0084$ & $.7484\pm.0021$ &
$+.0001\pm.0080$ & $-.0032\pm.0060$ \\
 & TUEV  & $.4813\pm.0192$ & $.4720\pm.0216$ & $.4763\pm.0052$ &
$-.0044\pm.0171$ & $-.0093\pm.0406$ \\
\midrule
LaBraM & FACED  & $.1499\pm.0003$ & $.1406\pm.0049$ & $.1940\pm.0024$ &
$-.0533\pm.0070$ & $-.0093\pm.0047$ \\
 & ISRUC$^{\dagger}$  & $.6354\pm.0019$ & $.7322\pm.0002$ & $.6977\pm.0059$ &
$+.0345\pm.0058$ & $+.0968\pm.0018$ \\
 & TUEV  & $.3483^{\ddagger}$ & $.3676\pm.0260$ & $.4329\pm.0010$ &
$-.0653\pm.0250$ & $+.0193\pm.0260$ \\
\bottomrule
\end{tabular}
\end{table}

\begin{figure}[!htbp]
\centering
\includegraphics[width=0.92\linewidth]{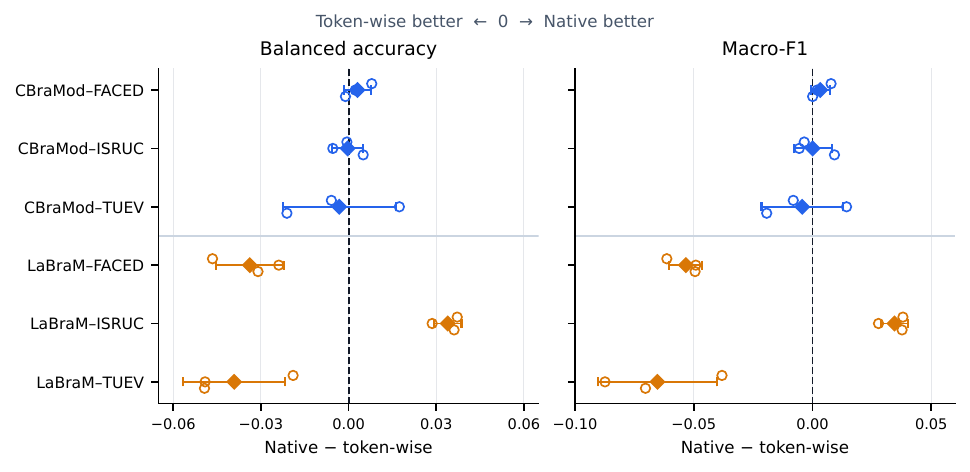}
\caption{Seed-level RQ1 effects for the six primary comparisons. Hollow points show
the three paired seed effects, diamonds show their means, and
horizontal intervals show across-seed standard deviations. Positive values
favor the native adapter. LaBraM--ISRUC uses the frozen-sequence protocol;
Section~\ref{sec:boundary-sequence} reports the sequence-trained follow-up.}
\label{fig:rq1-seed-effects}
\end{figure}

For RQ1, CBraMod's
mean effects range from $-0.004$ to $+0.003$ across the two primary metrics
and are smaller than the corresponding across-seed standard deviations.
Neither adapter shows a consistent advantage across the three CBraMod
settings. For LaBraM, the effects are
larger and setting-dependent. The native adapter is weaker on FACED by
0.034 balanced-accuracy points and 0.053 macro-F1 points and on TUEV by
0.039 and 0.065 points, respectively. The primary ISRUC comparison favors
the native adapter by 3.40 and 3.45 percentage points, but freezes a newly
initialized sequence encoder. Training that encoder reduces these differences
to 0.23 and 0.22 points, respectively (Section~\ref{sec:boundary-sequence}).
ISRUC therefore shows protocol sensitivity rather than a general native-adapter
advantage.

Under the primary frozen-sequence protocol, all three LaBraM seeds favor
the native adapter on ISRUC; all three favor the token-wise control on FACED
and TUEV (Figure~\ref{fig:rq1-seed-effects}).

Validation trajectories provide additional context for these selected-checkpoint
comparisons (Appendix Figure~\ref{fig:training-dynamics}). Appendix~\ref{app:selector-sensitivity}
examines checkpoint-selection sensitivity separately.

\subsection{Probe improvement and adapter preference (RQ2)}

For RQ2, the native adapter improves both metrics on CBraMod--FACED and
LaBraM--ISRUC. LaBraM--TUEV also has positive mean effects, with greater seed
variation. The LaBraM--FACED effect is near zero in balanced accuracy and
negative in macro-F1; CBraMod--ISRUC and CBraMod--TUEV have negative means.
Appendix Table~\ref{tab:rq2-supporting} reports additional native-adapter
versus probe comparisons on PhysioNet-MI and SEED-V.

\begin{figure}[!ht]
\centering
\includegraphics[width=0.93\linewidth]{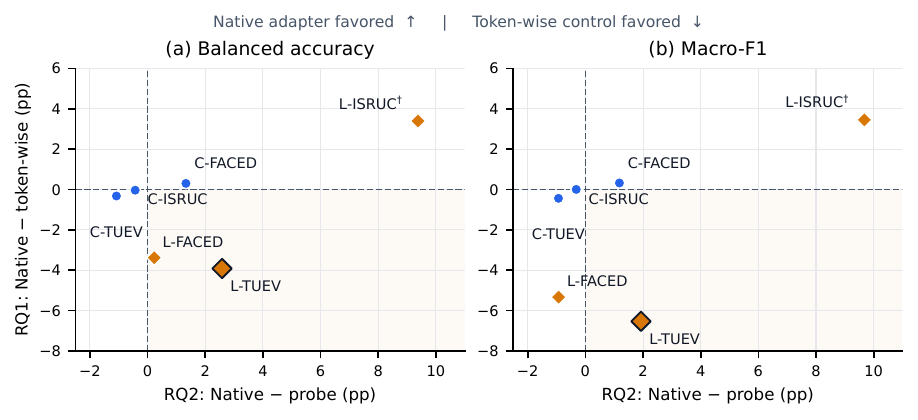}
\caption{Mean RQ2 (native $-$ probe; horizontal) and RQ1 (native $-$ token-wise;
vertical) effects in percentage points across three paired seeds. The shaded
lower-right region indicates mean improvement over the probe but lower
performance than the token-wise control; outlined LaBraM--TUEV points illustrate
this for both metrics. Uncertainty is reported in Table~\ref{tab:primary-effects}
and Figure~\ref{fig:rq1-seed-effects}. C/L denote CBraMod/LaBraM.
$\dagger$ LaBraM--ISRUC freezes a newly initialized sequence encoder;
Section~\ref{sec:boundary-sequence} reports the follow-up. LaBraM--TUEV RQ2
uses the reevaluated common probe reference (Table~\ref{tab:primary-effects}).}
\label{fig:rq1-rq2-quadrants}
\end{figure}

Improvement over the probe does not establish which adapter design is
preferable. On LaBraM--TUEV, the native adapter has positive mean effects over
the probe but negative effects against the token-wise control on both metrics
(Figure~\ref{fig:rq1-rq2-quadrants}). LaBraM--ISRUC favors the native adapter
under both contrasts in the primary frozen-sequence protocol, while CBraMod's
RQ1 effects remain near zero.

The absolute scores also matter for interpretation. On LaBraM--FACED, balanced
accuracy ranges from .159 to .195 for the probe, native adapter, and token-wise
control, while LoRA and Upper-2 reach .282 and .297 on the same split (Appendix
Table~\ref{tab:labram-context}). LoRA and Upper-2 provide broader performance
context rather than additional capacity-matched adapter comparisons.

\subsection{Secondary trainable-backbone comparison (RQ3)}
\label{sec:boundary-trainability}

RQ3 compares full fine-tuning with and without the native adapter, updating
the pretrained encoder and classifier in both conditions (Appendix
Table~\ref{tab:rq3-effects}). Adding the adapter improves mean balanced
accuracy and macro-F1 on CBraMod--TUEV, CBraMod--SEED-V, and
LaBraM--PhysioNet-MI, but reduces both metrics on CBraMod--FACED.

On TUEV, the balanced-accuracy gain is $1.27\pm3.70$ percentage points and
the macro-F1 gain is $2.49\pm8.58$ points, with standard deviations exceeding
the mean gains. Adding the native adapter therefore provides gains in some
settings, without a consistent benefit across the four comparisons.
Unlike RQ1, these comparisons do not include a token-wise control.

\FloatBarrier
\Needspace{10\baselineskip}
\begin{samepage}
\section{Robustness and Diagnostic Analyses}
\label{sec:boundary-conditions}

Targeted follow-up experiments test sensitivity to the branch operator and
downstream sequence-encoder training. The FACED/ISRUC branch-preserving MLP
and LaBraM--ISRUC sequence-encoder follow-ups used configurations and contrasts
fixed before those runs. Validation trajectories, selector comparisons,
class-level analyses, and residual-update measurements diagnose existing
primary results.
\end{samepage}

\subsection{Branch-preserving MLP control}
\label{sec:boundary-operator}

To test the sensitivity of the ranking to the branch operator, we replace
native attention with a branch-local multilayer perceptron (MLP). This control
preserves branch extraction but transforms tokens independently, with widths
adjusted to match parameter counts.

For LaBraM, against the branch-preserving MLP, the native adapter remains weaker on FACED
and TUEV and stronger on ISRUC under the primary frozen-sequence protocol,
on all three seeds for both metrics (Table~\ref{tab:labram-operator-audit}).
The two MLP controls have similar mean performance. FACED and ISRUC
provide initialization-matched tests; historical TUEV branch controls use
seed-specific adapter initialization. CBraMod--TUEV effects remain small
relative to seed variability (Appendix~\ref{app:operator-audit}).

\begin{table}[!ht]
\centering
\caption{LaBraM branch-preserving MLP comparison at validation-$\kappa$-selected checkpoints (mean $\pm$ SD, three paired seeds). Native: native adapter; Branch: branch-preserving MLP; Token: token-wise control. BA: balanced accuracy; F1: macro-F1. $\dagger$ TUEV Branch uses seed-specific adapter initialization; all other conditions use adapter seed 12345. ISRUC uses a newly initialized frozen sequence encoder (Table~\ref{tab:primary-configs}).}
\label{tab:labram-operator-audit}
\vspace{\baselineskip}
\tablefont
\setlength{\tabcolsep}{3pt}
\begin{tabular}{@{}llccccc@{}}
\toprule
Dataset & Metric & Native & Branch & Token & Native $-$ Branch & Branch $-$ Token \\
\midrule
FACED & BA & $.1613\pm.0097$ & $.1996\pm.0034$ & $.1951\pm.0028$ & $-.0384\pm.0113$ & $+.0046\pm.0055$ \\
FACED & F1 & $.1406\pm.0049$ & $.1947\pm.0032$ & $.1940\pm.0024$ & $-.0540\pm.0080$ & $+.0007\pm.0027$ \\
ISRUC & BA & $.7453\pm.0003$ & $.7136\pm.0049$ & $.7114\pm.0044$ & $+.0317\pm.0051$ & $+.0023\pm.0065$ \\
ISRUC & F1 & $.7322\pm.0002$ & $.7003\pm.0055$ & $.6977\pm.0059$ & $+.0319\pm.0057$ & $+.0026\pm.0072$ \\
TUEV$^{\dagger}$ & BA & $.3916\pm.0182$ & $.4311\pm.0033$ & $.4307\pm.0010$ & $-.0395\pm.0208$ & $+.0003\pm.0043$ \\
TUEV$^{\dagger}$ & F1 & $.3676\pm.0260$ & $.4336\pm.0030$ & $.4329\pm.0010$ & $-.0660\pm.0288$ & $+.0007\pm.0039$ \\
\bottomrule
\end{tabular}
\end{table}

\subsection{Sequence-encoder training on ISRUC}
\label{sec:boundary-sequence}

The primary LaBraM--ISRUC protocol freezes a newly initialized downstream
sequence encoder. We repeat all four conditions with this encoder trained
jointly with the classifier and any adapter, while keeping the pretrained
epoch encoder frozen. The follow-up uses the same three seeds, subject
splits, 30-epoch budget, and validation-$\kappa$ selector; details and
seed-level results are in Appendix~\ref{app:isruc-sequence-sensitivity}.

\begin{table}[!htbp]
\centering
\caption{LaBraM--ISRUC sequence-encoder sensitivity (mean $\pm$ SD, three paired seeds). Both protocols freeze the pretrained epoch encoder and train the classifier and any adapter. The baseline has no adapter; only the follow-up trains the newly initialized sequence encoder. BA: balanced accuracy; F1: macro-F1.}
\label{tab:isruc-sequence-sensitivity}
\vspace{\baselineskip}
\tablefont
\setlength{\tabcolsep}{3pt}
\begin{tabular}{@{}lcccc@{}}
\toprule
 & \multicolumn{2}{c}{Frozen sequence encoder} & \multicolumn{2}{c}{Trainable sequence encoder} \\
Condition & BA & F1 & BA & F1 \\
\midrule
No-adapter baseline & $.6515\pm.0021$ & $.6354\pm.0019$ & $.7581\pm.0016$ & $.7502\pm.0030$ \\
Native adapter & $.7453\pm.0003$ & $.7322\pm.0002$ & $.7713\pm.0036$ & $.7641\pm.0037$ \\
Token-wise control & $.7114\pm.0044$ & $.6977\pm.0059$ & $.7689\pm.0065$ & $.7619\pm.0078$ \\
Branch-preserving MLP & $.7136\pm.0049$ & $.7003\pm.0055$ & $.7676\pm.0041$ & $.7611\pm.0048$ \\
\bottomrule
\end{tabular}
\end{table}

The sequence-trained protocol improves all four conditions
(Table~\ref{tab:isruc-sequence-sensitivity}). All three adapters outperform
the no-adapter baseline on every seed for both metrics. The native adapter has
the highest means, but its advantage over the token-wise control is substantially
smaller than in the primary frozen-sequence comparison; macro-F1 changes sign
on one seed. The native adapter exceeds
the branch-preserving MLP on all three seeds, by 0.37 BA and 0.30 F1 points
on average. The magnitude and consistency of the ISRUC preference are therefore
sensitive to the sequence-encoder training protocol.

\subsection{Class-specific TUEV behavior}
\label{sec:boundary-classes}

The LaBraM difference is concentrated in generalized periodic epileptiform discharges
(GPED) and eye movement (EYEM), where native F1 is lower by 0.114
and 0.190, respectively; CBraMod effects are comparatively small
(Figure~\ref{fig:tuev-class-effects}). Absolute values are in
Appendix~\ref{app:tuev-class-values}.
Macro-F1 gives
each class equal weight, so GPED and EYEM contribute substantially to the gap
despite their different test-set supports. The class-level results locate the performance differences without
identifying their causes.

\begin{figure}[H]
\centering
\includegraphics[width=0.90\linewidth]{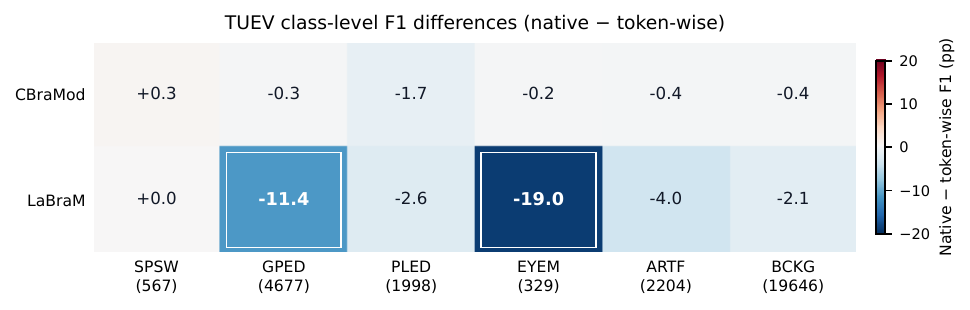}
\caption{TUEV class-level F1 differences (native $-$ token-wise; percentage
points), averaged over three seeds at validation-$\kappa$-selected checkpoints.
Outlines mark LaBraM's largest mean deficits. Parentheses give test-set class
support; both backbones share a zero-centered scale.
Appendix Table~\ref{tab:tuev-class-effects} reports across-seed variation.}
\label{fig:tuev-class-effects}
\end{figure}
\FloatBarrier

\Needspace{8\baselineskip}
\section{Discussion}
\label{sec:discussion}

\subsection{Implications for downstream adaptation}
\label{sec:discussion-main}

The matched comparisons distinguish the value of adding adaptation capacity
from the choice of adapter. A native adapter can improve over a frozen-backbone
probe while remaining weaker than the token-wise control, as on LaBraM--TUEV.
The branch-preserving MLP provides complementary evidence: retaining the branch
layout without attention yields similar mean performance to the token-wise
control.

The ISRUC follow-up shows that adapter comparisons are sensitive to downstream
training conditions. Under the sequence-trained protocol, the native adapter's
advantage becomes much smaller (Section~\ref{sec:boundary-sequence}). The larger
frozen-sequence effect therefore provides protocol-specific evidence rather
than a general preference for the native adapter.

CBraMod's small effects remain unresolved rather than establishing equivalence.
Both adapters produce measurable residual updates, so their similar performance
is not simply due to either remaining inactive
(Appendix~\ref{app:cbramod-residuals}). Pretrained feature interactions and
downstream classifier mixing may affect the value of additional adapter-level
interactions, but these comparisons do not isolate their contributions, or
those of adapter operators and optimization.

Two follow-up comparisons could clarify these unresolved contributions.
A parameter-matched axis-mixing MLP would retain cross-token interaction
without attention, complementing the branch-preserving control. On ISRUC,
varying sequence-encoder trainability while holding dropout behavior and
other settings fixed would help distinguish their contributions to the
protocol sensitivity.

\subsection{Limitations}
\label{sec:discussion-limitations}

We evaluate two backbones across six primary backbone--dataset settings and
three training seeds, with adapters placed after the pretrained encoder.
These runs do not cover all sources of initialization uncertainty. Adapter
pairs share optimization settings; other insertion points and separate tuning
remain untested.

RQ1 compares complete attention-based and token-wise MLP adapter designs, so
it does not isolate axis mixing. The native adapter follows each backbone's
axes and feature partitions, but we do not test alternative arrangements or
whether CBraMod's final feature halves retain distinct channel and temporal
roles. Adapter implementations, classifier heads, and optimization also differ
across backbones. CBraMod's flattened-feature heads, for example, already mix
information across the representation, which may reduce the benefit of further
adapter-level mixing. Differences across backbones therefore characterize the
evaluated pipelines and cannot be attributed to backbone identity alone.

The primary LaBraM--ISRUC protocol freezes a newly initialized sequence
encoder; the follow-up trains it while retaining the frozen pretrained
epoch encoder (Appendix~\ref{app:isruc-sequence-sensitivity}). This sensitivity
analysis changes both sequence-encoder training and its dropout behavior. CBraMod--TUEV pretraining
overlap is unresolved. LaBraM--TUEV uses a common probe reference with
unverified independent training replication
(Appendix~\ref{app:rq2-seed-effects}). Available
held-out predictions limit selector-sensitivity analysis
(Appendix~\ref{app:selector-sensitivity}).

\subsection{Conclusion}
\label{sec:discussion-future}

In this study, we examined whether downstream adaptation should follow the channel and temporal
structure of pretrained EEG features. Across six primary matched comparisons,
the native adapter shows no consistent advantage over the token-wise control.
The preferred design varies across backbone--task settings and downstream
training protocols, with the sleep-staging comparison particularly sensitive
to sequence-encoder training.
Pretrained feature organization provides a starting point for adapter design.
Evaluating backbone-informed adaptation requires comparisons that account for
both trainable capacity and downstream training conditions; improvement over
a frozen-backbone probe alone does not establish a preference for the structured
design.

\FloatBarrier

\bibliographystyle{tmlr}
\bibliography{references/tmlr}

\appendix
\newpage
\section{Primary Per-Seed RQ1 Results}
\label{app:primary-seed-results}

Table~\ref{tab:primary-seed-rq1} reports the held-out test values underlying
the primary RQ1 aggregate effects. All entries use the checkpoint selected by
validation Cohen's $\kappa$. Effects are native minus token-wise and are
paired within seed.

\begin{table}[h]
\centering
\caption{Per-seed primary RQ1 test results at validation-$\kappa$-selected
checkpoints. BA denotes balanced accuracy and F1 macro-F1; $\Delta$ is native
minus token-wise performance in 0--1 score units.}
\label{tab:primary-seed-rq1}
\vspace{\baselineskip}
\tablefont
\setlength{\tabcolsep}{2.5pt}
\begin{tabular}{@{}llrcccccc@{}}
\hline
\textbf{Backbone} & \textbf{Dataset} & \textbf{Seed} & \textbf{Native BA} & \textbf{Token-wise BA} & \textbf{$\Delta$BA} & \textbf{Native F1} & \textbf{Token-wise F1} & \textbf{$\Delta$F1} \\
\hline
CBraMod & FACED & 42 & .3952 & .3873 & $+.0079$ & .3951 & .3873 & $+.0078$ \\
CBraMod & FACED & 1024 & .3875 & .3853 & $+.0023$ & .3879 & .3860 & $+.0019$ \\
CBraMod & FACED & 3407 & .3902 & .3913 & $-.0011$ & .3891 & .3891 & $+.0001$ \\
CBraMod & ISRUC & 42 & .7541 & .7547 & $-.0006$ & .7472 & .7506 & $-.0035$ \\
CBraMod & ISRUC & 1024 & .7446 & .7500 & $-.0054$ & .7408 & .7464 & $-.0056$ \\
CBraMod & ISRUC & 3407 & .7570 & .7520 & $+.0050$ & .7575 & .7482 & $+.0092$ \\
CBraMod & TUEV & 42 & .4426 & .4485 & $-.0059$ & .4638 & .4719 & $-.0081$ \\
CBraMod & TUEV & 1024 & .4805 & .4631 & $+.0174$ & .4964 & .4821 & $+.0143$ \\
CBraMod & TUEV & 3407 & .4413 & .4624 & $-.0211$ & .4557 & .4749 & $-.0193$ \\
LaBraM & TUEV & 42 & .4127 & .4317 & $-.0190$ & .3959 & .4340 & $-.0381$ \\
LaBraM & TUEV & 1024 & .3806 & .4297 & $-.0491$ & .3448 & .4322 & $-.0874$ \\
LaBraM & TUEV & 3407 & .3815 & .4308 & $-.0493$ & .3622 & .4325 & $-.0703$ \\
LaBraM & FACED & 42 & .1515 & .1981 & $-.0466$ & .1352 & .1965 & $-.0614$ \\
LaBraM & FACED & 1024 & .1708 & .1947 & $-.0239$ & .1444 & .1936 & $-.0491$ \\
LaBraM & FACED & 3407 & .1614 & .1924 & $-.0310$ & .1423 & .1918 & $-.0495$ \\
LaBraM & ISRUC & 42 & .7456 & .7085 & $+.0372$ & .7323 & .6943 & $+.0381$ \\
LaBraM & ISRUC & 1024 & .7451 & .7165 & $+.0286$ & .7323 & .7045 & $+.0278$ \\
LaBraM & ISRUC & 3407 & .7453 & .7092 & $+.0361$ & .7320 & .6943 & $+.0377$ \\
\hline
\end{tabular}
\end{table}

\FloatBarrier

\section{Experimental Details}
\label{app:evidence-scope}

This appendix gives dataset construction and training configurations for the
comparisons in Section~\ref{sec:experimental-protocol}.

We applied the primary comparison criteria retrospectively to existing runs.
For the subsequent LaBraM FACED/ISRUC matched comparisons and CBraMod--ISRUC
optimization-matched controls, we fixed configurations before training.
We also fixed configurations and contrasts before the FACED/ISRUC
branch-preserving controls and LaBraM--ISRUC sequence-encoder follow-up.

We use the public CBraMod checkpoint.\footnote{\url{https://huggingface.co/weighting666/CBraMod}.
SHA-256 prefix: \texttt{0792cb808c14}.}
The original study reports standard and TUEV-excluded models
\citep{wang2025cbramod}, but the release model card does not identify
which variant these weights represent.

\subsection{Dataset preprocessing and input construction}
\label{app:preprocessing}

We follow the CBraMod preprocessing protocol \citep{wang2025cbramod}.
Within each paired comparison, both adapter conditions use the same
preprocessed inputs and splits. Table~\ref{tab:dataset-construction}
gives input shapes, split definitions, and example counts.

\begin{table}[H]
\centering
\caption{Model inputs and dataset splits. Counts denote ten-second segments
for FACED, 20-epoch sequences for ISRUC, five-second records for TUEV,
one-second windows for SEED-V, and four-second epochs for PhysioNet-MI.
ISRUC input shapes include the sequence dimension.}
\label{tab:dataset-construction}
\vspace{\baselineskip}
\tablefont
\setlength{\tabcolsep}{3pt}
\begin{tabular}{@{}llp{0.35\linewidth}r@{}}
\toprule
\textbf{Dataset} & \textbf{Input shape} & \textbf{Split definition} & \textbf{Train / Val. / Test} \\
\midrule
FACED & $(32,10,200)$ & Subjects 000--079 / 080--099 / 100--122 & 6,720 / 1,680 / 1,932 \\
ISRUC & $(20,6,30,200)$ & Subjects 1--80 / 81--90 / 91--100 & 3,559 / 468 / 435 \\
TUEV & $(16,5,200)$ & Record/prefix split; original evaluation set used for testing & 68,712 / 15,220 / 29,421 \\
SEED-V & $(62,1,200)$ & Within-subject trials 0--4 / 5--9 / 10--14 & 34,432 / 42,960 / 40,352 \\
PhysioNet-MI & $(64,4,200)$ & 70 / 19 / 20 subjects & 6,300 / 1,734 / 1,803 \\
\bottomrule
\end{tabular}
\end{table}

\paragraph{FACED.}
We use the processed 32-channel recordings for nine-class emotion recognition
\citep{chen2023faced}. We resample each processed trial to 6,000 samples
and divide it into three non-overlapping ten-second segments.
Each segment contains ten
one-second patches with 200 samples per channel. Subjects are disjoint across
the training, validation, and test sets.

\paragraph{ISRUC.}
We use six bipolar channels for five-class sleep staging
\citep{khalighi2016isruc}, in the order
\texttt{F3-A2, C3-A2, O1-A2, F4-A1, C4-A1, O2-A1}.
The preprocessing script takes 200 Hz signals as input, applies a
0.3--35 Hz bandpass filter and a 50 Hz notch filter, and constructs
30-second epochs containing 6,000 samples per channel.
We group consecutive epochs into non-overlapping sequences of 20 and discard
incomplete epochs and sequences at the end of each recording.
Each epoch retains its sleep-stage label.

\paragraph{TUEV.}
We use six-class EEG event labels \citep{obeid2016tuh,harati2015tuev}.
We resample recordings to 200 Hz, apply a 0.3--75 Hz bandpass filter and a
60 Hz notch filter, and construct 16 differential channels in the order
\texttt{FP1-F7, F7-T3, T3-T5, T5-O1, FP2-F8, F8-T4, T4-T6, T6-O2},
followed by
\texttt{FP1-F3, F3-C3, C3-P3, P3-O1, FP2-F4, F4-C4, C4-P4, P4-O2}.
Each example spans five seconds, including two seconds of context on either
side of the annotated event. We divide sorted training-record prefixes into
80\% training and 20\% validation groups and use the original evaluation set
for testing. The input files do not specify a TUEV release version.

\paragraph{SEED-V.}
We use 62 EEG channels for five-class emotion recognition \citep{liu2022seedv}.
We resample recordings to 200 Hz, apply a 0.3--75 Hz bandpass filter, and
segment each annotated trial into non-overlapping one-second windows.
We split trials within each subject and session into training, validation,
and test sets before constructing the windows. Each example contains one
temporal patch, giving the $S=1$ boundary condition.

\paragraph{PhysioNet-MI.}
We use the four motor-imagery classes from 64-channel recordings
\citep{goldberger2000physiobank,schalk2009eegmmidb}.
We interpolate channels marked as bad, apply an average reference,
a 0.3 Hz high-pass filter, and a 60 Hz notch filter, then resample to 200 Hz.
We extract four-second epochs from the task annotations, exclude rest events,
and retain 800 samples per channel. Subjects are disjoint across splits.

\paragraph{Backbone input formatting.}
We divide amplitudes by 100 for FACED, TUEV, SEED-V, and PhysioNet-MI and
leave ISRUC amplitudes unchanged. Each temporal patch contains 200 samples.
Both backbones receive the same preprocessed examples and splits.
LaBraM maps FACED and PhysioNet-MI channels to their model positions;
for TUEV, it uses sequential positional slots 1--16.

\subsection{Primary experimental configurations}
\label{app:primary-configs}

Table~\ref{tab:primary-configs} gives the primary RQ1 configurations,
shared across seeds 42, 1024, and 3407. Adapter parameter counts
exclude the shared downstream head. The epoch budgets apply to both adapter
conditions and the probe; Figure~\ref{fig:training-dynamics} shows validation
trajectories over these budgets.

All primary conditions use AdamW with base weight decay $0.05$ and label
smoothing $0.1$. Adapters use dropout $0$ and residual multiplier $\gamma=1$.
We select checkpoints by validation Cohen's $\kappa$.

\begin{table}[h]
\centering
\caption{Primary frozen-backbone configurations, shared across the three seeds.
$\dagger$ The newly initialized sequence encoder is frozen in all three conditions.}
\label{tab:primary-configs}
\vspace{\baselineskip}
\tablefont
\setlength{\tabcolsep}{2pt}
\begin{tabular}{@{}lllrrrrrrr@{}}
\toprule
\multicolumn{3}{c}{\textbf{Setting}} & \multicolumn{2}{c}{\textbf{Architecture}} & \multicolumn{4}{c}{\textbf{Optimization}} & \textbf{Capacity} \\
\cmidrule(lr){1-3}\cmidrule(lr){4-5}\cmidrule(lr){6-9}\cmidrule(lr){10-10}
\textbf{Backbone} & \textbf{Dataset} & \textbf{Condition} & \textbf{$r$} & \textbf{Heads} & \textbf{Adapter LR} & \textbf{Head LR} & \textbf{Batch} & \textbf{Training epochs} & \textbf{Adapter params} \\
\midrule
CBraMod & FACED & Native & 64 & 4 & $5\times10^{-4}$ & $5\times10^{-4}$ & 64 & 50 & 59,610 \\
CBraMod & FACED & Token-wise & 148 & -- & $5\times10^{-4}$ & $5\times10^{-4}$ & 64 & 50 & 59,949 \\
CBraMod & ISRUC & Native & 64 & 4 & $2\times10^{-5}$ & $3.536\times10^{-4}$ & 8 & 20 & 59,610 \\
CBraMod & ISRUC & Token-wise & 148 & -- & $2\times10^{-5}$ & $3.536\times10^{-4}$ & 8 & 20 & 59,949 \\
CBraMod & TUEV & Native & 64 & 4 & $5\times10^{-5}$ & $2\times10^{-4}$ & 8 & 20 & 59,610 \\
CBraMod & TUEV & Token-wise & 148 & -- & $5\times10^{-5}$ & $2\times10^{-4}$ & 8 & 20 & 59,949 \\
LaBraM & FACED & Native & 64 & 4 & $7\times10^{-4}$ & $7\times10^{-4}$ & 32 & 80 & 85,810 \\
LaBraM & FACED & Token-wise & 213 & -- & $7\times10^{-4}$ & $7\times10^{-4}$ & 32 & 80 & 86,014 \\
LaBraM & ISRUC$^{\dagger}$ & Native & 64 & 4 & $2\times10^{-4}$ & $2\times10^{-4}$ & 16 & 30 & 85,810 \\
LaBraM & ISRUC$^{\dagger}$ & Token-wise & 213 & -- & $2\times10^{-4}$ & $2\times10^{-4}$ & 16 & 30 & 86,014 \\
LaBraM & TUEV & Native & 64 & 4 & $5\times10^{-4}$ & $5\times10^{-4}$ & 64 & 15 & 85,810 \\
LaBraM & TUEV & Token-wise & 213 & -- & $5\times10^{-4}$ & $5\times10^{-4}$ & 64 & 15 & 86,014 \\
\bottomrule
\end{tabular}
\end{table}

All adaptation modules initialize the residual scale at $\alpha=0.01$.
LaBraM uses low-rank native attention branches and a full token-wise
bottleneck control. Output projections are zero-initialized
for CBraMod and not zero-initialized for LaBraM.

Here $r$ denotes the bottleneck dimension. CBraMod uses model/classifier
dropout $0.1$ and a per-iteration cosine schedule with minimum learning
rate $10^{-6}$. For LaBraM, the model dropout is $0.0$,
stochastic-depth rates are $0.1$ for FACED/ISRUC and $0.2$ for TUEV, and
the schedules use ten, one, and five warm-up epochs for FACED, ISRUC, and TUEV,
respectively. Frozen-backbone evaluation disables stochastic depth.

\begin{table}[!htbp]
\centering
\caption{Downstream heads in the six primary frozen-backbone comparisons.
Counts include trainable downstream-head parameters and exclude adapters.
Head architecture and trainability are shared by native, token-wise, and probe
conditions within each comparison. Linear-layer dimensions are input-to-output.}
\label{tab:primary-heads}
\vspace{\baselineskip}
\tablefont
\setlength{\tabcolsep}{3pt}
\begin{tabular}{@{}llp{0.57\linewidth}r@{}}
\toprule
\textbf{Backbone} & \textbf{Dataset} & \textbf{Downstream head} & \textbf{Trainable params} \\
\midrule
CBraMod & FACED & Flatten; $64{,}000\to2{,}000\to200\to9$ MLP;
ELU and dropout $0.1$ after the two hidden layers & 128,404,009 \\
CBraMod & ISRUC & Per-epoch flatten; $36{,}000\to512$ projection and GELU;
one Transformer layer ($d=512$, 4 heads, feed-forward width 2,048,
dropout $0.1$); $512\to5$ classifier & 21,587,461 \\
CBraMod & TUEV & Flatten; $16{,}000\to1{,}000\to200\to6$ MLP;
ELU and dropout $0.1$ after the two hidden layers & 16,202,406 \\
LaBraM & FACED & Mean patch pooling; frozen LayerNorm; $200\to9$ classifier & 1,809 \\
LaBraM & ISRUC & Per-epoch mean pooling; frozen LayerNorm;
newly initialized, frozen Transformer layer ($d=200$, 4 heads,
feed-forward width 800); $200\to5$ classifier & 1,005 \\
LaBraM & TUEV & Mean patch pooling; frozen LayerNorm; $200\to6$ classifier & 1,206 \\
\bottomrule
\end{tabular}
\par\smallskip
\begin{minipage}{\linewidth}
Both ISRUC sequence Transformers use GELU and pre-layer normalization.
The LaBraM sequence encoder contains 482,600 parameters. In the
sequence-trained follow-up, it and the classifier contribute 483,605
trainable parameters, with sequence dropout $0.1$ active during training.
\end{minipage}
\end{table}

The CBraMod--ISRUC residual scales and remaining adapter parameters use
weight decay $0.05$ in both conditions, for all three seeds.

In the primary LaBraM--ISRUC comparison, the newly initialized sequence
encoder remains frozen; training updates only the $200$-to-$5$ classifier
(1,005 parameters) and any adapter. CBraMod trains its sequence encoder.
Appendix~\ref{app:isruc-sequence-sensitivity} reports the LaBraM follow-up
with a trainable sequence encoder.

\FloatBarrier
\begin{minipage}{\linewidth}
\subsection{TUEV class-level values}
\label{app:tuev-class-values}

Table~\ref{tab:tuev-class-effects} gives the absolute class-level values
underlying Figure~\ref{fig:tuev-class-effects}.

\begin{table}[H]
\centering
\caption{TUEV class-level results at validation-$\kappa$-selected checkpoints.
$N$ denotes test examples and R recall. Recall and F1 are means over three
seeds; $\Delta$ is native minus token-wise, with sample standard deviations
in parentheses. Scores and effects are in 0--1 units.}
\label{tab:tuev-class-effects}
\vspace{\baselineskip}
\tablefont
\setlength{\tabcolsep}{2.5pt}
\begin{tabular}{@{}llrrrrrrr@{}}
\hline
\textbf{Backbone} & \textbf{Class} & \textbf{$N$} & \textbf{Native R} & \textbf{Token-wise R} & \textbf{$\Delta$R} & \textbf{Native F1} & \textbf{Token-wise F1} & \textbf{$\Delta$F1} \\
\hline
CBraMod & SPSW & 567 & .039 & .039 & $+.000\;(.000)$ & .065 & .061 & $+.003\;(.003)$ \\
CBraMod & GPED & 4677 & .353 & .356 & $-.003\;(.012)$ & .465 & .468 & $-.003\;(.015)$ \\
CBraMod & PLED & 1998 & .310 & .323 & $-.014\;(.040)$ & .369 & .385 & $-.017\;(.033)$ \\
CBraMod & EYEM & 329 & .585 & .585 & $+.000\;(.053)$ & .697 & .700 & $-.002\;(.046)$ \\
CBraMod & ARTF & 2204 & .542 & .539 & $+.002\;(.014)$ & .364 & .368 & $-.004\;(.008)$ \\
CBraMod & BCKG & 19646 & .901 & .906 & $-.005\;(.008)$ & .872 & .876 & $-.004\;(.004)$ \\
LaBraM & SPSW & 567 & .000 & .000 & $+.000\;(.000)$ & .000 & .000 & $+.000\;(.000)$ \\
LaBraM & GPED & 4677 & .345 & .466 & $-.121\;(.143)$ & .427 & .541 & $-.114\;(.110)$ \\
LaBraM & PLED & 1998 & .469 & .461 & $+.008\;(.046)$ & .360 & .386 & $-.026\;(.034)$ \\
LaBraM & EYEM & 329 & .077 & .185 & $-.108\;(.094)$ & .118 & .308 & $-.190\;(.134)$ \\
LaBraM & ARTF & 2204 & .582 & .562 & $+.020\;(.060)$ & .420 & .460 & $-.040\;(.051)$ \\
LaBraM & BCKG & 19646 & .877 & .910 & $-.033\;(.006)$ & .881 & .902 & $-.021\;(.007)$ \\
\hline
\end{tabular}
\end{table}
\end{minipage}
\par

\FloatBarrier

\section{Supporting Results and Robustness Analyses}
\label{app:supporting-results}

This appendix contains contextual absolute-performance comparisons, the
secondary trainability analysis, checkpoint-selection comparisons, supporting
dataset results, and the ISRUC sequence-encoder follow-up.

\subsection{Contextual adaptation comparisons}

Table~\ref{tab:method-context} compares held-out test performance across
adaptation methods within each backbone--dataset setting. We select
checkpoints by validation Cohen's $\kappa$.

The generic bottleneck adapter uses the same token-wise residual MLP form as
the token-wise control: layer normalization, a down-projection, GELU, and an
up-projection, with a learned residual scale. Its bottleneck width is fixed
at 64 (26,265 trainable adapter parameters), whereas the primary token-wise
control uses width 148 for CBraMod or 213 for LaBraM to approximately match
the native adapter's capacity. The Generic columns therefore report a
smaller-capacity version of the same architecture.

\begin{table}[h]
\centering
\caption{Contextual held-out test performance for settings with broader reference
coverage. Entries are mean $\pm$ standard deviation over three seeds at the
validation-selected checkpoint; dashes indicate unavailable conditions.
Generic denotes a generic bottleneck adapter, Upper denotes upper-layer
fine-tuning, and Dense denotes full fine-tuning. $\ddagger$ LaBraM--TUEV
uses a common probe reference
(Appendix~\ref{app:rq2-seed-effects}).}
\label{tab:method-context}
\vspace{\baselineskip}
\tablefont
\setlength{\tabcolsep}{2pt}
\begin{tabular}{@{}llccccccc@{}}
\hline
\multicolumn{9}{@{}l@{}}{\textbf{(a) Balanced accuracy}} \\
\hline
\textbf{Backbone} & \textbf{Dataset} & \textbf{Probe} & \textbf{Native} & \textbf{Token-wise} & \textbf{Generic} & \textbf{LoRA} & \textbf{Upper} & \textbf{Dense} \\
\hline
CBraMod & FACED & .378$\pm$.002 & .391$\pm$.004 & .388$\pm$.003 & .390$\pm$.003 & .417$\pm$.007 & .403$\pm$.005 & .572$\pm$.011 \\
CBraMod & ISRUC & .756$\pm$.003 & .752$\pm$.006 & .752$\pm$.002 & -- & -- & -- & -- \\
CBraMod & TUEV & .466$\pm$.031 & .455$\pm$.022 & .458$\pm$.008 & .461$\pm$.012 & .555$\pm$.009 & .557$\pm$.031 & .620$\pm$.024 \\
LaBraM & TUEV & .3658$^{\ddagger}$ & .392$\pm$.018 & .431$\pm$.001 & .437$\pm$.013 & .572$\pm$.023 & .541$\pm$.020 & -- \\
\hline
\end{tabular}
\par\vspace{4pt}
\begin{tabular}{@{}llccccccc@{}}
\hline
\multicolumn{9}{@{}l@{}}{\textbf{(b) Macro-F1}} \\
\hline
\textbf{Backbone} & \textbf{Dataset} & \textbf{Probe} & \textbf{Native} & \textbf{Token-wise} & \textbf{Generic} & \textbf{LoRA} & \textbf{Upper} & \textbf{Dense} \\
\hline
CBraMod & FACED & .379$\pm$.002 & .391$\pm$.004 & .387$\pm$.002 & .389$\pm$.004 & .418$\pm$.008 & .404$\pm$.005 & .579$\pm$.012 \\
CBraMod & ISRUC & .752$\pm$.005 & .748$\pm$.008 & .748$\pm$.002 & -- & -- & -- & -- \\
CBraMod & TUEV & .481$\pm$.019 & .472$\pm$.022 & .476$\pm$.005 & .474$\pm$.007 & .527$\pm$.006 & .518$\pm$.020 & .565$\pm$.037 \\
LaBraM & TUEV & .3483$^{\ddagger}$ & .368$\pm$.026 & .433$\pm$.001 & .432$\pm$.012 & .540$\pm$.004 & .525$\pm$.017 & -- \\
\hline
\end{tabular}
\end{table}

\FloatBarrier
\subsection{LaBraM cross-dataset context}
\label{app:labram-context}

Table~\ref{tab:labram-context} reports LaBraM--FACED and LaBraM--ISRUC
matched runs. Values are validation-kappa-selected test results averaged over the
three seeds. The three primary conditions (probe, native, and token-wise)
use a frozen backbone and the same dataset-specific training protocol.

\begin{table}[h]
\centering
\caption{LaBraM test results on FACED and ISRUC (mean $\pm$ standard
deviation over three seeds). BA denotes balanced accuracy, C+P channel-plus-patch
adaptation, and $b$ the token-wise bottleneck width. Upper-2 tunes the last two
backbone blocks. The ISRUC primary probe, native, and token-wise conditions
freeze a newly initialized sequence encoder. Paired effects are reported in
the main text.}
\label{tab:labram-context}
\vspace{\baselineskip}
\tablefont
\setlength{\tabcolsep}{2.5pt}
\begin{tabular}{@{}lllcccl@{}}
\hline
\textbf{Dataset} & \textbf{Condition} & \textbf{Regime} & \textbf{BA} & \textbf{Macro-F1} & \textbf{$n$} & \textbf{Qualification} \\
\hline
FACED & Frozen probe & Frozen & .159$\pm$.000 & .150$\pm$.000 & 3 & Probe \\
FACED & Native C+P, rank 64 & Frozen & .161$\pm$.010 & .141$\pm$.005 & 3 & Primary \\
FACED & Token-wise $b=213$ & Frozen & .195$\pm$.003 & .194$\pm$.002 & 3 & Primary control \\
FACED & LoRA & LoRA & .282$\pm$.032 & .280$\pm$.032 & 3 & Context \\
FACED & Upper-2 & Upper layer & .297$\pm$.001 & .297$\pm$.002 & 3 & Context \\
ISRUC & Frozen probe & Frozen & .652$\pm$.002 & .635$\pm$.002 & 3 & Probe \\
ISRUC & Native C+P, rank 64 & Frozen & .745$\pm$.000 & .732$\pm$.000 & 3 & Primary \\
ISRUC & Token-wise $b=213$ & Frozen & .711$\pm$.004 & .698$\pm$.006 & 3 & Primary control \\
ISRUC & Native C+P, rank 32 & Trainable & .792$\pm$.006 & .785$\pm$.008 & 3 & Context \\
ISRUC & LoRA & LoRA & .692$\pm$.004 & .677$\pm$.006 & 3 & Context \\
ISRUC & Upper-2 & Upper layer & .794$\pm$.005 & .785$\pm$.004 & 3 & Context \\
\hline
\end{tabular}
\end{table}

\FloatBarrier
\subsection{Branch-preserving MLP comparison}
\label{app:operator-audit}

The branch-preserving MLP preserves native channel/patch branch extraction and
replaces attention with independent token transformations. Its widths are 147
for CBraMod and 106 for LaBraM, compared with native rank 64. The realized
parameter mismatches are 0.144\% and 0.471\%, respectively.
Table~\ref{tab:labram-operator-audit} reports the three-way LaBraM comparison;
Table~\ref{tab:operator-new-seeds} gives the six new FACED/ISRUC seed results.

FACED/ISRUC branch controls use the primary learning rates, batches,
schedules, 80/30-epoch budgets, adapter seed 12345, and classifier trainability.
ISRUC retains the newly initialized frozen sequence encoder.
LaBraM--TUEV branch controls use adapter seeds
30042, 31024, and 33407; its native and ordinary token-wise conditions use 12345.

\begin{table}[!ht]
\centering
\caption{New branch-preserving MLP controls on LaBraM FACED/ISRUC. B denotes Branch, A denotes the native adapter, and T denotes the token-wise control. Each row uses its validation-$\kappa$-selected checkpoint.}
\label{tab:operator-new-seeds}
\vspace{\baselineskip}
\tablefont
\setlength{\tabcolsep}{3pt}
\begin{tabular}{@{}llrrrrrr@{}}
\toprule
Dataset & Seed & B BA & A $-$ B BA & B $-$ T BA & B F1 & A $-$ B F1 & B $-$ T F1 \\
\midrule
FACED & 42 & $.1990$ & $-.0475$ & $+.0009$ & $.1978$ & $-.0626$ & $+.0013$ \\
FACED & 1024 & $.1966$ & $-.0258$ & $+.0019$ & $.1913$ & $-.0469$ & $-.0022$ \\
FACED & 3407 & $.2033$ & $-.0419$ & $+.0109$ & $.1949$ & $-.0526$ & $+.0031$ \\
ISRUC & 42 & $.7089$ & $+.0367$ & $+.0004$ & $.6942$ & $+.0381$ & $-.00003$ \\
ISRUC & 1024 & $.7133$ & $+.0318$ & $-.0032$ & $.7016$ & $+.0306$ & $-.0028$ \\
ISRUC & 3407 & $.7187$ & $+.0266$ & $+.0095$ & $.7050$ & $+.0270$ & $+.0107$ \\
\bottomrule
\end{tabular}
\end{table}

\begin{table}[h]
\centering
\caption{Branch-preserving MLP comparison on TUEV. Entries are mean $\pm$
standard deviation over three seeds. The effect is native adapter minus
branch-preserving MLP; negative values favor the branch-preserving MLP.
MLP denotes the branch-preserving MLP control.}
\label{tab:operator-audit}
\vspace{\baselineskip}
\tablefont
\setlength{\tabcolsep}{2pt}
\begin{tabular}{@{}lrrrrrrcc@{}}
\hline
\textbf{Backbone} & \textbf{Native BA} & \textbf{MLP BA} & \textbf{$\Delta$BA} &
\textbf{Native F1} & \textbf{MLP F1} & \textbf{$\Delta$F1} &
\textbf{BA MLP $>$} & \textbf{F1 MLP $>$} \\
\hline
CBraMod & .4548$\pm$.0223 & .4637$\pm$.0168 & $-.0089\pm$.0315 &
 .4720$\pm$.0216 & .4781$\pm$.0063 & $-.0061\pm$.0235 & 2/3 & 2/3 \\
LaBraM & .3916$\pm$.0182 & .4311$\pm$.0033 & $-.0395\pm$.0208 &
 .3676$\pm$.0260 & .4336$\pm$.0030 & $-.0660\pm$.0288 & 3/3 & 3/3 \\
\hline
\end{tabular}
\end{table}

On TUEV, the branch-preserving MLP outperforms the native adapter on all three
LaBraM seeds and on two of three CBraMod seeds for both metrics.

\FloatBarrier
\Needspace{6\baselineskip}
\subsection{Secondary trainability analysis}

The four full-fine-tuning comparisons train all backbone and classifier
weights, with an adapter added only in the native condition.
Table~\ref{tab:rq3-configs} gives the optimizer settings.

\begin{table}[!htbp]
\centering
\caption{Configurations for the four RQ3 full-fine-tuning pairs, shared across
seeds 42, 1024, and 3407. Native rank is 64; C/P denote channel/patch axes.
Learning rates are peak values.
Both conditions use AdamW, cosine schedules, label smoothing 0.1, and
validation-$\kappa$ selection. WD denotes base weight decay.}
\label{tab:rq3-configs}
\vspace{\baselineskip}
\tablefont
\setlength{\tabcolsep}{3pt}
\begin{tabular}{@{}llcrrccc@{}}
\toprule
Backbone & Dataset & Axes & Epochs & Batch & Backbone/upper LR & Head/adapter LR & WD \\
\midrule
CBraMod & FACED & C+P & 50 & 64 & $10^{-4}/10^{-4}$ & $5\!\times\!10^{-4}/5\!\times\!10^{-4}$ & .05 \\
CBraMod & TUEV & C+P & 20 & 8 & $10^{-4}/2\!\times\!10^{-4}$ & $2\!\times\!10^{-4}/2\!\times\!10^{-4}$ & .05 \\
CBraMod & SEED-V & C & 40 & 32 & $10^{-4}/10^{-4}$ & $10^{-4}/10^{-4}$ & .03 \\
LaBraM & PhysioNet-MI & C+P & 40 & 64 & $5\!\times\!10^{-4}/5\!\times\!10^{-4}$ & $5\!\times\!10^{-4}/5\!\times\!10^{-4}$ & .05 \\
\bottomrule
\end{tabular}
\end{table}

CBraMod upper-group rates apply to the last two backbone blocks; FACED uses
one common backbone rate.
CBraMod--TUEV exempts the adapter scalar from weight decay. LaBraM uses
five warmup epochs and exempts biases and one-dimensional parameters from
weight decay.

\begin{table}[h]
\centering
\caption{RQ3 incremental native-adapter effects under frozen and
trainable-backbone regimes (mean $\pm$ SD over three seeds). Frozen effects
are native minus probe; trainable effects are full fine-tuning with the native
adapter minus full fine-tuning without the adapter. Dashes indicate
unavailable comparisons. $\ddagger$ LaBraM--TUEV frozen effects use the
common probe reference (Appendix~\ref{app:rq2-seed-effects}).
BA denotes balanced accuracy and F1 macro-F1; effects are in 0--1 score units.}
\label{tab:rq3-effects}
\vspace{\baselineskip}
\tablefont
\begin{tabular}{@{}llcccc@{}}
\hline
\textbf{Backbone} & \textbf{Dataset} & \multicolumn{2}{c}{\textbf{Frozen}} & \multicolumn{2}{c}{\textbf{Trainable}} \\
 & & \textbf{BA} & \textbf{F1} & \textbf{BA} & \textbf{F1} \\
\hline
CBraMod & FACED & $+0.0133\pm0.0061$ & $+0.0118\pm0.0061$ & $-0.0064\pm0.0083$ & $-0.0073\pm0.0091$ \\
CBraMod & ISRUC & $-0.0043\pm0.0050$ & $-0.0032\pm0.0060$ & -- & -- \\
CBraMod & TUEV & $-0.0108\pm0.0511$ & $-0.0093\pm0.0406$ & $+0.0127\pm0.0370$ & $+0.0249\pm0.0858$ \\
LaBraM & TUEV$^{\ddagger}$ & $+0.0258\pm0.0182$ & $+0.0193\pm0.0260$ & -- & -- \\
LaBraM & PhysioNet-MI & $+0.0139\pm0.0057$ & $+0.0054\pm0.0085$ & $+0.0159\pm0.0074$ & $+0.0164\pm0.0064$ \\
CBraMod & SEED-V & $-0.0069\pm0.0036$ & $-0.0074\pm0.0042$ & $+0.0077\pm0.0068$ & $+0.0085\pm0.0068$ \\
\hline
\end{tabular}
\end{table}

\FloatBarrier
\begin{minipage}{\linewidth}
\subsection{Checkpoint-selection differences}
\label{app:selector-sensitivity}

Table~\ref{tab:selector-sensitivity} compares checkpoint epochs selected by
validation Cohen's $\kappa$ and balanced accuracy across three seeds.

\begin{table}[H]
\centering
\caption{Checkpoint-epoch differences between validation Cohen's $\kappa$
and balanced-accuracy selection. ``Same'' counts matching selections among
three seeds; median and mean shifts are absolute epoch differences.}
\label{tab:selector-sensitivity}
\vspace{\baselineskip}
\tablefont
\begin{tabular}{@{}llccc@{}}
\hline
\textbf{Setting} & \textbf{Condition} & \textbf{Same} & \textbf{Median shift} & \textbf{Mean shift} \\
\hline
CBraMod--FACED & Native & 2/3 & 0.0 & 4.0 \\
CBraMod--FACED & Token-wise & 2/3 & 0.0 & 1.7 \\
CBraMod--TUEV & Native & 0/3 & 3.0 & 4.0 \\
CBraMod--TUEV & Token-wise & 0/3 & 7.0 & 6.0 \\
LaBraM--TUEV & Native & 0/3 & 6.0 & 5.3 \\
LaBraM--TUEV & Token-wise & 0/3 & 3.0 & 3.0 \\
\hline
\end{tabular}
\end{table}
\end{minipage}
\par

\FloatBarrier
\subsection{Supporting and boundary RQ2 comparisons}

\begin{table}[H]
\centering
\caption{Frozen-probe effects by backbone and dataset for supporting and boundary conditions.
Effects are native minus probe (mean $\pm$ SD over three seeds, in 0--1 score
units). BA denotes balanced accuracy and F1 macro-F1; BA $+$ and F1 $+$ count
seeds with positive effects. Effects for the primary settings are in Table~\ref{tab:primary-effects}.}
\label{tab:rq2-supporting}
\vspace{\baselineskip}
\tablefont
\begin{tabular}{@{}llccccc@{}}
\hline
\textbf{Backbone} & \textbf{Dataset} & \textbf{BA} & \textbf{F1} & \textbf{BA $+$} & \textbf{F1 $+$} \\
\hline
CBraMod & PhysioNet-MI & $+0.0014\pm0.0102$ & $+0.0010\pm0.0087$ & 1/3 & 1/3 \\
LaBraM & PhysioNet-MI & $+0.0139\pm0.0057$ & $+0.0054\pm0.0085$ & 3/3 & 2/3 \\
CBraMod & SEED-V & $-0.0069\pm0.0036$ & $-0.0074\pm0.0042$ & 0/3 & 0/3 \\
\hline
\end{tabular}
\end{table}

\FloatBarrier
\subsection{Per-seed RQ2 effects}
\label{app:rq2-seed-effects}

Table~\ref{tab:rq2-seed-results} reports the numerical seed-level values for
the primary frozen-probe comparisons. The native adapter and probe are
evaluated at their respective validation-kappa-selected checkpoints; the
effect is native minus probe.

\begin{table}[h]
\centering
\caption{Per-seed RQ2 held-out test results for the primary settings. LaBraM--TUEV uses a common probe reference, as explained below.
RQ2 effects are native minus probe.}
\label{tab:rq2-seed-results}
\vspace{\baselineskip}
\tablefont
\setlength{\tabcolsep}{2pt}
\begin{tabular}{@{}llr rrr rrr@{}}
\hline
\textbf{Backbone} & \textbf{Dataset} & \textbf{Seed} & \multicolumn{3}{c}{\textbf{Balanced accuracy}} & \multicolumn{3}{c}{\textbf{Macro-F1}} \\
 & & & \textbf{Native} & \textbf{Probe} & \textbf{$\Delta$} & \textbf{Native} & \textbf{Probe} & \textbf{$\Delta$} \\
\hline
CBraMod & FACED & 42 & .3952 & .3751 & $+.0201$ & .3951 & .3762 & $+.0189$ \\
CBraMod & FACED & 1024 & .3875 & .3791 & $+.0085$ & .3879 & .3796 & $+.0083$ \\
CBraMod & FACED & 3407 & .3902 & .3790 & $+.0113$ & .3891 & .3808 & $+.0083$ \\
CBraMod & ISRUC & 42 & .7541 & .7543 & $-.0002$ & .7472 & .7466 & $+.0006$ \\
CBraMod & ISRUC & 1024 & .7446 & .7544 & $-.0098$ & .7408 & .7509 & $-.0101$ \\
CBraMod & ISRUC & 3407 & .7570 & .7600 & $-.0030$ & .7575 & .7575 & $+.0000$ \\
CBraMod & TUEV & 42 & .4426 & .4986 & $-.0561$ & .4638 & .4927 & $-.0289$ \\
CBraMod & TUEV & 1024 & .4805 & .4360 & $+.0445$ & .4964 & .4591 & $+.0373$ \\
CBraMod & TUEV & 3407 & .4413 & .4623 & $-.0210$ & .4557 & .4921 & $-.0364$ \\
LaBraM & TUEV & 42 & .4127 & .3658 & $+.0469$ & .3959 & .3483 & $+.0476$ \\
LaBraM & TUEV & 1024 & .3806 & .3658 & $+.0149$ & .3448 & .3483 & $-.0035$ \\
LaBraM & TUEV & 3407 & .3815 & .3658 & $+.0157$ & .3622 & .3483 & $+.0138$ \\
LaBraM & FACED & 42 & .1515 & .1587 & $-.0072$ & .1352 & .1497 & $-.0146$ \\
LaBraM & FACED & 1024 & .1708 & .1593 & $+.0115$ & .1444 & .1502 & $-.0058$ \\
LaBraM & FACED & 3407 & .1614 & .1587 & $+.0027$ & .1423 & .1497 & $-.0074$ \\
LaBraM & ISRUC & 42 & .7456 & .6539 & $+.0917$ & .7323 & .6375 & $+.0948$ \\
LaBraM & ISRUC & 1024 & .7451 & .6502 & $+.0949$ & .7323 & .6346 & $+.0977$ \\
LaBraM & ISRUC & 3407 & .7453 & .6504 & $+.0948$ & .7320 & .6341 & $+.0979$ \\
\hline
\end{tabular}
\end{table}

\FloatBarrier

Separate evaluation of three saved LaBraM--TUEV probe checkpoints reproduces
identical classification scores. We therefore report a common probe reference;
independent training replication remains unverified, and these scores do not
establish low seed variability.

\subsection{LaBraM--ISRUC sequence-encoder sensitivity}
\label{app:isruc-sequence-sensitivity}

We repeat the native, token-wise, branch-preserving MLP, and no-adapter
conditions with the newly initialized sequence encoder trained jointly with
the classifier and any adapter. The pretrained epoch encoder remains frozen
in evaluation mode. We retain the subject splits and inputs in
Appendix~\ref{app:preprocessing}, the three training seeds, and the
30-epoch budget. Training the sequence encoder also activates its dropout
($0.1$), so this follow-up tests the sequence-training protocol.

We pair sequence-encoder and classifier initialization within each seed and
initialize adapters with seed 12345 independently of the training RNG.
Native and token-wise adapter counts match Table~\ref{tab:primary-configs};
the branch-preserving MLP contains 86,214 parameters. Matching excludes the
shared sequence encoder and classifier (Table~\ref{tab:primary-heads}).

We retain the LaBraM--ISRUC batch size, learning rate, warmup, and
regularization settings in Appendix~\ref{app:primary-configs}, applying the
same peak learning rate $2\times10^{-4}$ to every trainable group.
AdamW exempts biases and one-dimensional parameters from weight decay;
the cosine schedule has minimum learning rate $10^{-6}$.
We select the first checkpoint attaining the maximum validation Cohen's
$\kappa$ and compute test BA and macro-F1 from its confusion matrix.
Tables~\ref{tab:isruc-sequence-seeds} and~\ref{tab:isruc-sequence-effects}
report all 12 runs and their paired effects.

\begin{table}[!htbp]
\centering
\caption{LaBraM--ISRUC follow-up test results with a trainable sequence encoder. Epoch is the validation-$\kappa$-selected training epoch (one-based). BA: balanced accuracy; F1: macro-F1.}
\label{tab:isruc-sequence-seeds}
\vspace{\baselineskip}
\tablefont
\setlength{\tabcolsep}{5pt}
\begin{tabular}{@{}lrrrr@{}}
\toprule
Condition & Seed & Epoch & BA & F1 \\
\midrule
No-adapter baseline & 42 & 25 & 0.7580 & 0.7477 \\
No-adapter baseline & 1024 & 23 & 0.7598 & 0.7536 \\
No-adapter baseline & 3407 & 23 & 0.7567 & 0.7493 \\
Native adapter & 42 & 23 & 0.7672 & 0.7598 \\
Native adapter & 1024 & 23 & 0.7739 & 0.7660 \\
Native adapter & 3407 & 26 & 0.7726 & 0.7665 \\
Token-wise control & 42 & 27 & 0.7614 & 0.7529 \\
Token-wise control & 1024 & 23 & 0.7728 & 0.7659 \\
Token-wise control & 3407 & 23 & 0.7725 & 0.7669 \\
Branch-preserving MLP & 42 & 26 & 0.7633 & 0.7558 \\
Branch-preserving MLP & 1024 & 23 & 0.7714 & 0.7650 \\
Branch-preserving MLP & 3407 & 23 & 0.7680 & 0.7625 \\
\bottomrule
\end{tabular}
\end{table}

\begin{table}[!htbp]
\centering
\caption{Paired effects with a trainable LaBraM--ISRUC sequence encoder. Entries are mean $\pm$ sample SD in percentage points; positive counts describe the three observed seeds. Contrasts were fixed before the follow-up runs.}
\label{tab:isruc-sequence-effects}
\vspace{\baselineskip}
\tablefont
\setlength{\tabcolsep}{4pt}
\begin{tabular}{@{}lcccc@{}}
\toprule
Contrast & BA & BA $+$ & Macro-F1 & F1 $+$ \\
\midrule
Native $-$ Token-wise & $+0.234\pm0.306$ & 3/3 & $+0.218\pm0.414$ & 2/3 \\
Native $-$ Branch & $+0.368\pm0.103$ & 3/3 & $+0.299\pm0.174$ & 3/3 \\
Branch $-$ Token-wise & $-0.134\pm0.320$ & 1/3 & $-0.080\pm0.366$ & 1/3 \\
Native $-$ Baseline & $+1.312\pm0.349$ & 3/3 & $+1.392\pm0.283$ & 3/3 \\
\bottomrule
\end{tabular}
\end{table}

\FloatBarrier

\subsection{CBraMod residual-update diagnostic}
\label{app:cbramod-residuals}

We load the nine primary native-adapter checkpoints and their nine token-wise
counterparts, selected by validation Cohen's $\kappa$, and measure residual
updates in evaluation mode without retraining. Both conditions use the same
validation splits and summary procedure. For each validation feature tensor $H$, we compute
$\|\Delta H\|_F/\|H\|_F$, where $\Delta H$ includes the learned branch scales
and residual multiplier $\gamma=1$. The full validation splits contain 1,680
FACED examples, 9,360 ISRUC epochs from 468 sequences, and 15,220 TUEV examples.
Table~\ref{tab:cbramod-residuals} reports learned scales and per-example ratio
summaries.

\begin{table}[!htbp]
\centering
\caption{CBraMod residual updates for nine primary native--token-wise pairs
at their validation-$\kappa$-selected checkpoints. Learned residual scales are
$\alpha_C$ and $\alpha_S$ for the native channel and temporal-patch branches,
and $\alpha_T$ for the token-wise control; dashes indicate inapplicable scales.
Ratio mean and SD summarize $\|\Delta H\|_F/\|H\|_F$ across validation examples,
with ISRUC measured per sleep epoch. These are feature-update magnitudes,
not performance or causal effects.}
\label{tab:cbramod-residuals}
\vspace{\baselineskip}
\tablefont
\setlength{\tabcolsep}{4pt}
\begin{tabular}{@{}lrlrrrrr@{}}
\toprule
\textbf{Dataset} & \textbf{Seed} & \textbf{Condition} & \textbf{$\alpha_C$} & \textbf{$\alpha_S$} & \textbf{$\alpha_T$} & \textbf{Ratio mean} & \textbf{Ratio SD} \\
\midrule
FACED & 42 & Native & 0.0308 & 0.0279 & -- & 0.6723 & 0.0433 \\
FACED & 42 & Token-wise & -- & -- & 0.0376 & 0.7350 & 0.0458 \\
FACED & 1024 & Native & 0.0304 & 0.0291 & -- & 0.6841 & 0.0416 \\
FACED & 1024 & Token-wise & -- & -- & 0.0368 & 0.7278 & 0.0440 \\
FACED & 3407 & Native & 0.0307 & 0.0281 & -- & 0.6680 & 0.0426 \\
FACED & 3407 & Token-wise & -- & -- & 0.0390 & 0.7290 & 0.0435 \\
ISRUC & 42 & Native & 0.0354 & 0.0386 & -- & 0.1076 & 0.0412 \\
ISRUC & 42 & Token-wise & -- & -- & 0.0369 & 0.0686 & 0.0283 \\
ISRUC & 1024 & Native & 0.0389 & 0.0423 & -- & 0.1431 & 0.0557 \\
ISRUC & 1024 & Token-wise & -- & -- & 0.0415 & 0.0885 & 0.0365 \\
ISRUC & 3407 & Native & 0.0367 & 0.0351 & -- & 0.1663 & 0.0645 \\
ISRUC & 3407 & Token-wise & -- & -- & 0.0416 & 0.0990 & 0.0402 \\
TUEV & 42 & Native & 0.0590 & 0.0642 & -- & 0.5645 & 0.0767 \\
TUEV & 42 & Token-wise & -- & -- & 0.1032 & 0.6250 & 0.0797 \\
TUEV & 1024 & Native & 0.0404 & 0.0425 & -- & 0.5371 & 0.0793 \\
TUEV & 1024 & Token-wise & -- & -- & 0.0628 & 0.5824 & 0.0831 \\
TUEV & 3407 & Native & 0.0391 & 0.0417 & -- & 0.5680 & 0.0784 \\
TUEV & 3407 & Token-wise & -- & -- & 0.0728 & 0.6096 & 0.0840 \\
\bottomrule
\end{tabular}
\end{table}

\FloatBarrier

Both designs produce non-negligible residual updates; their magnitudes alone
do not explain the performance differences between adapters.

\subsection{Validation trajectories}
\label{sec:boundary-trajectories}

\begin{figure}[!htbp]
\centering
\includegraphics[width=0.88\linewidth]{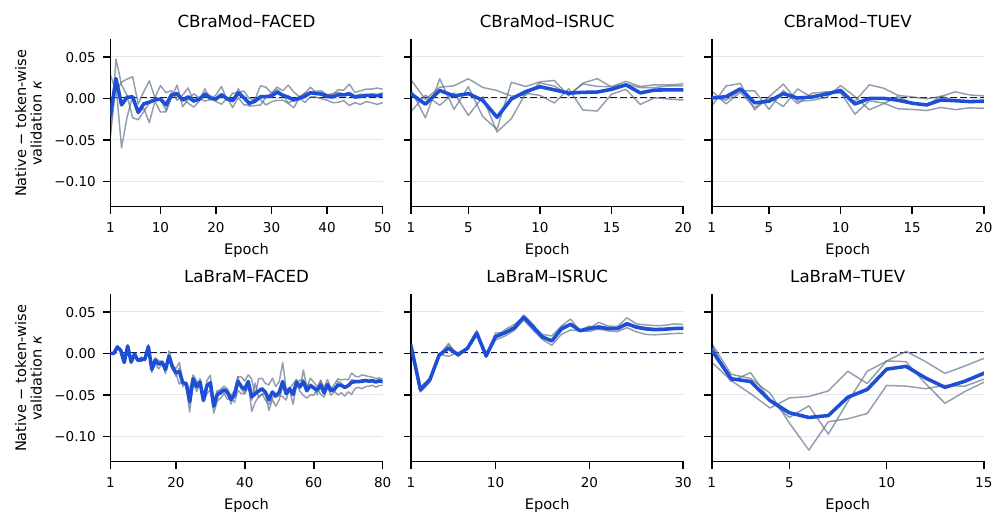}
\caption{Validation trajectories for the six primary comparisons. Lines show native
minus token-wise validation Cohen's $\kappa$ for three seeds (gray) and their
mean (blue); positive values favor native. Each panel shows its own recorded
training budget, matching Table~\ref{tab:primary-configs}. LaBraM--ISRUC
uses the primary frozen-sequence protocol; Section~\ref{sec:boundary-sequence}
reports sensitivity to sequence-encoder training.}
\label{fig:training-dynamics}
\end{figure}

Figure~\ref{fig:training-dynamics} reports validation Cohen's $\kappa$
trajectories for the six primary comparisons. Appendix~\ref{app:selector-sensitivity}
compares the checkpoint epochs selected by validation Cohen's $\kappa$
and balanced accuracy.

\FloatBarrier

\end{document}